\documentclass[11pt]{article}

\usepackage[preprint]{acl}

\usepackage{times}
\usepackage{latexsym}

\usepackage[T1]{fontenc}

\usepackage[utf8]{inputenc}
\usepackage{amsmath,amssymb,amsthm}

\usepackage{booktabs}
\usepackage{colortbl}
\usepackage{tabularx}
\definecolor{oursblue}{RGB}{232,243,252}
\usepackage{multirow}
\usepackage{algorithm}
\usepackage{algpseudocode}
\usepackage{tcolorbox}
\usepackage{bookmark}
\tcbuselibrary{breakable,listings,skins}
\newlength{\promptindent}
\newtcblisting{agentprompt}[1]{
  enhanced, listing only, width=\columnwidth,
  title={#1},
  fonttitle=\bfseries\small, coltitle=white,
  colbacktitle=black, colframe=black, colback=black!4,
  boxrule=0.6pt, arc=1.5mm,
  left=1.5mm, right=1.5mm, top=1mm, bottom=1mm,
  before skip=6pt, after skip=6pt,
  listing options={basicstyle=\scriptsize\ttfamily, columns=fullflexible,
    keepspaces=true, breaklines=true, breakatwhitespace=true,
    breakindent=\promptindent,
    showstringspaces=false, upquote=true, aboveskip=0pt, belowskip=0pt}
}
\usepackage{microtype}

\usepackage{inconsolata}

\usepackage{graphicx}
\usepackage{capt-of} 
\title{OGDA: Ontology-Grounded Data Agents for Reliable Analysis over Heterogeneous Data Lakes} 

\author{Boyi Zhang\textsuperscript{1}, Hong Ting Tsang\textsuperscript{1}, Yisen Gao\textsuperscript{1},
  \textbf{Qiao Xiao\textsuperscript{1}, LI Yufei\textsuperscript{1}} \\
  \textbf{Haoyu Huang\textsuperscript{1}, Zhongwei Xie\textsuperscript{1}, Jiaxin Bai\textsuperscript{2}, Yangqiu Song\textsuperscript{1}} \\
  \textsuperscript{1}The Hong Kong University of Science and Technology \\
  \textsuperscript{2}Hong Kong Baptist University \\
  {\small\texttt{\{bzhangdd,httsangaj,ygaodi,qxiaoc,ylivm,hhuangcp,zxiebk\}@connect.ust.hk}} \\
  {\small\texttt{baijiaxin@hkbu.edu.hk}, \texttt{yqsong@cse.ust.hk}}}

\begin{document}
\maketitle

\begin{abstract}
Full-pipeline data analysis over real-world data lakes requires LLM agents to locate and integrate information across large collections of tables with different schemas. Existing data agents follow task-specific exploration, which exhausts the context window in massive lakes. As the context fills, the agent's reasoning degrades, leading it to locate and process incorrect data files. These errors propagate into the final answer. To overcome these limitations, we introduce Ontology-Grounded Data Agent (\ours{}), which utilizes a compact ontology to reduce exploration overhead and constrain data processing. \ours{} constructs the ontology through agentic hypothesis testing. At each step, a proposer agent induces schemas, concepts, and relations from raw sources, and a verifier checks each proposal against the underlying data. During task execution, the ontology serves as a representation of the data lake for the agent to explore. Every data-processing action is constrained by a task graph, which the agent can extend only after observing the semantic meaning of the data. Across KramaBench and DiscoveryBench with three LLM backbones, \ours{} outperforms the strongest baselines by up to 10.82 points. These results suggest that explicitly modeling the data lake is a key step toward reliable end-to-end data agents.
\end{abstract}
\section{Introduction}
\label{sec:introduction}

\begin{figure}[!t]
    \centering
    \includegraphics[width=\columnwidth]{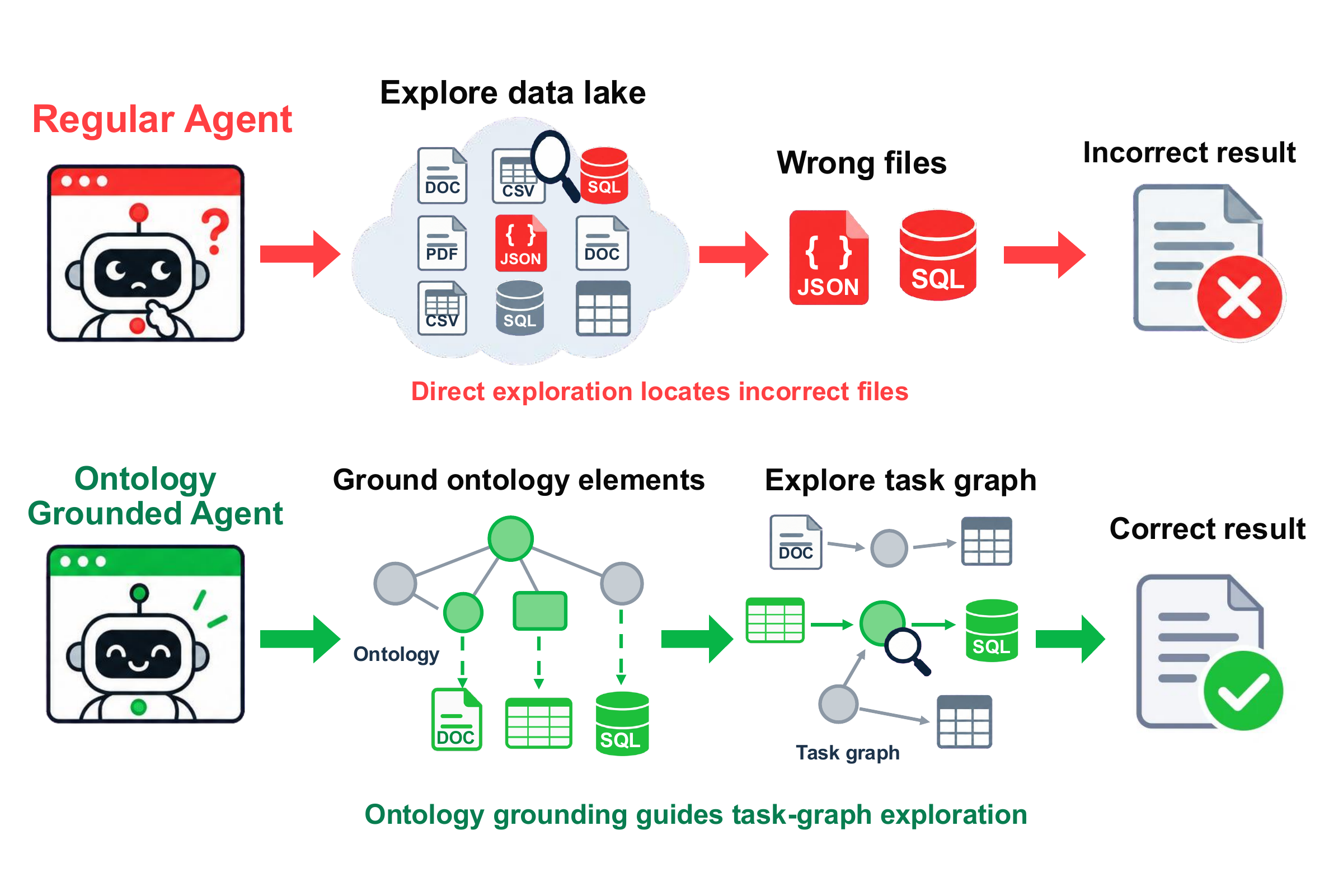}
    \caption{Direct exploration can select incorrect files and propagate source-selection errors into the result. \ours{} grounds relevant ontology elements in an explicit task graph that guides exploration of the corresponding sources.}
    \label{fig:agent-comparison}
\end{figure}

Data analysis answers a question end-to-end from raw data, spanning data preparation, integration, and computation. Traditional approaches decompose this process into separately designed stages, including data discovery, cleaning, integration, and computation~\citep{8509315,wang-etal-2020-rat,yu-etal-2018-spider}. Each stage assumes a predefined problem space with fixed objectives and a known target structure. Thus, it must be redesigned by hand whenever the data or question changes.

LLMs have shown strong reasoning abilities required for data analysis, from interpreting tabular data to generating analysis code that answers complex questions~\citep{cheng-etal-2023-gpt,3618408.3619164}. Motivated by such capabilities, many works utilize LLMs to remove much of the manual design by translating natural-language questions directly into executable queries or analysis code~\citep{hong-etal-2024-knowledge,10.14778/3641204.3641221,pourreza2023dinsql}. Furthermore, LLM agents extend this capability to multi-step tasks. By interleaving reasoning with actions, agents gather information, revise plans, and carry out tasks through successive interactions with an environment~\citep{yao2023react}. Building on this paradigm, data analysis agents automate the data analysis pipeline, from locating relevant files to iteratively refining analysis solutions~\citep{nam2026dsstar,hong-etal-2025-data,you-etal-2025-datawiseagent}. Their design implicitly assumes that the data relevant to a task can be found and understood by exploring sources within the agent's context.

However, this assumption does not hold in real-world data lakes. A lake typically mixes various relational tables with no shared schema describing what they contain or how their fields relate~\citep{lei2025spider,lai2026kramabench}. As illustrated in Figure~\ref{fig:agent-comparison}, this exploration often exhausts the agent's context window before the task reaches its final step. Moreover, evidence discovered early is likely to be buried under the growing exploration trace, causing hallucinations and error propagation~\citep{zhang2024how,liu-etal-2024-lost}. Existing context management methods mitigate context limitations by managing external memory, summarizing completed subgoals, or compressing interaction histories~\citep{hu-etal-2025-hiagent,liu-etal-2026-context,zhang-etal-2025-treerare}. While these approaches improve context efficiency, they do not explicitly model the semantic meaning of the underlying data or the relationships between sources. Data analysts confront the same problem when managing large, heterogeneous collections, and a common practice is to maintain a semantic layer over the data~\citep{10.1145/2882903.2899389,ijcai2018p777,10.1145/2882903.2903730}. A table's schema describes what each field inside the table represents. Building on this, an ontology of the data lake records how various fields connect through shared concepts, so that an analyst can locate and combine the data a question requires without inspecting every raw file. 

Inspired by this practice, grounding data agents in an ontology offers a promising way to guide analysis using the semantic knowledge underlying a data lake. However, realizing this approach presents two challenges. First, real-world data lakes rarely provide a predefined ontology. Second, merely supplying an ontology as additional context does not ensure that an agent will use it effectively. Without explicit constraints, the agent may still explore irrelevant files or perform computations without consulting the available semantic knowledge.

To address these challenges, we propose Ontology-Grounded Data Agent (\ours{})\footnote{Code is available at \url{https://github.com/HKUST-KnowComp/OGDA}.}, a framework that automatically constructs an ontology over a data lake and uses it to govern both the context and the actions of an LLM agent. First, \emph{agentic ontology construction} abstracts the lake into an ontology in which schemas group files holding the same kind of record and concepts align semantically equivalent fields across files. During construction, a data agent proposes each schema, concept, and edge as a hypothesis induced from exploration of the data lake. An independent verifier tests each hypothesis by designing and executing experiments over the lake. \emph{Experience distillation} captures lessons from this exploration and verification. These lessons complement the ontology with knowledge beyond its explicit elements. Second, \emph{ontology-grounded action} turns this representation into an explicit constraint on execution. The agent must commit the ontology elements relevant to the task to a task graph, and any action on data outside that graph is rejected. The ontology thus orchestrates both what enters the agent's context and which data the agent may act on, so the agent's context scales with the task rather than with the size of the lake.

In summary, our contributions are:
\begin{itemize}
    \item We introduce \ours{}, which orchestrates agent execution and context through an ontology-based task graph.
    \item  We develop a hypothesis-testing procedure that constructs structural and semantic representations from raw files and distills reusable experience from exploration.
    \item Our experiments demonstrate that \ours{} significantly outperforms existing methods across KramaBench~\citep{lai2026kramabench} and DiscoveryBench~\citep{majumder2025discoverybench} over three backbone models.
\end{itemize}

\begin{figure*}[t]
    \centering
    \includegraphics[width=\textwidth]{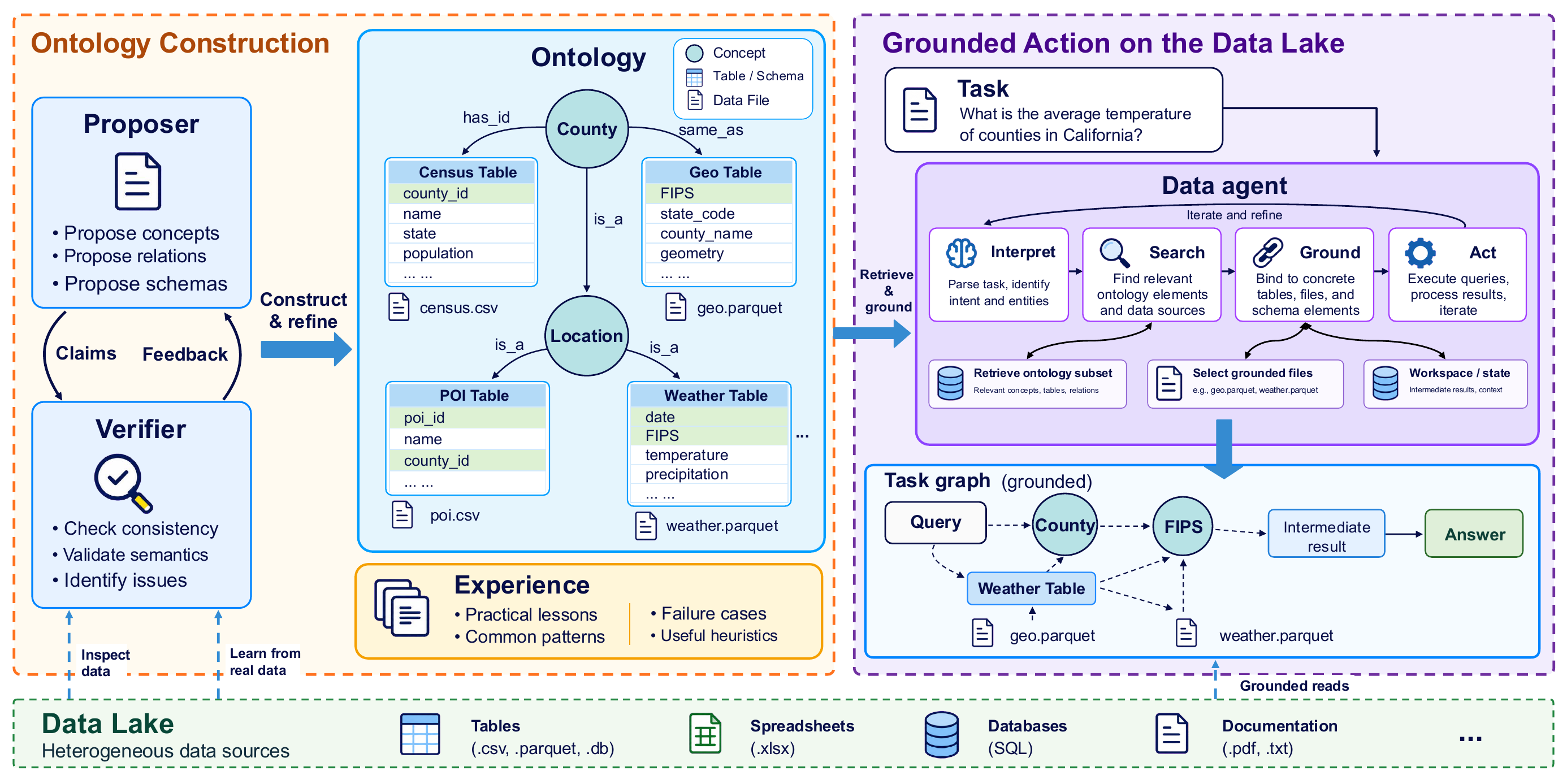}
    \caption{Overview of \ours{}. Left: a proposer and verifier construct and refine an ontology of the heterogeneous data lake through claims and feedback, while exploration yields reusable experience. Right: a data agent interprets a task, searches the ontology, grounds relevant elements in a task graph, and acts on the corresponding data. The task graph connects grounded files, intermediate results, and the final answer.}
    \label{fig:method-overview}
\end{figure*}
\section{Related Work}
\subsection{Data Agents}
Early LLM-based data systems answer questions over a given structured source. Text-to-SQL methods such as DIN-SQL~\citep{pourreza2023dinsql} and MAC-SQL~\citep{wang-etal-2025-mac} decompose a question, link its terms to the tables and columns of a database, and correct queries through feedback from either LLMs or the execution environment. StructGPT~\citep{jiang-etal-2023-structgpt} lets an LLM iteratively read evidence from tables, knowledge graphs, and databases through specialized interfaces. 
 
Data analysis agents extend this paradigm to full analyses over files, spanning data preparation, modeling, and interpretation. ReAct~\citep{yao2023react} interleaves reasoning with actions, while Reflexion~\citep{shinn2023reflexion} uses verbal feedback to improve subsequent attempts. For data science, DS-Agent~\citep{DS-Agent} reuses past successful solutions through case-based reasoning. Data Interpreter~\citep{hong-etal-2025-data} decomposes a task into a hierarchical graph of subtasks with stepwise verification, and DatawiseAgent~\citep{you-etal-2025-datawiseagent} organizes planning, incremental execution, and self-debugging around a computational notebook. DS-STAR~\citep{nam2026dsstar} targets heterogeneous file formats by first describing each data file and then iteratively planning, executing, and verifying analysis steps. These methods organize how an agent plans and carries out an analysis. \ours{} additionally represents cross-file correspondences in an ontology and requires the agent to inspect schema descriptions before grounding the relevant files in a task graph. This graph determines which raw data the agent may read.
 
\subsection{Semantic Layer Construction}
\label{sec:related-semantic-layer}
A \emph{semantic layer} maps the underlying data to domain concepts and their relations, so that queries can be posed over concepts rather than isolated files~\citep{ijcai2018p777}. Think-on-Graph~\citep{sun2024thinkongraph} and RoG~\citep{luo2024reasoning} let LLMs reason along paths in a knowledge graph, and OG-RAG~\citep{sharma-etal-2025-og} anchors retrieval in a domain ontology. These systems assume that the semantic layer already exists.
  
Constructing such a layer typically requires expert effort, and prior work automates it in several ways. Aurum~\citep{8509315} supports data discovery through relationships such as content and schema similarity across sources. LLM-based methods induce semantic layers from text~\citep{zhang-soh-2024-extract, bai-etal-2026-autoschemakg} or from database schemas and their associated questions~\citep{baek-etal-2025-knowledge}. EvoOntology~\citep{chong2026evoontologyselfevolvingontologylayer} constructs an ontology through workload-guided data probes and refines it using agent trajectories, accepting updates through paired evaluation. In contrast, \ours{} explores the lake independently of downstream tasks to extract implicit domain knowledge and verify these interpretations against the underlying sources before incorporating them into the ontology.

\section{Method}
\label{sec:method}

In this section, we first formalize the data analysis task and the ontology over a data lake in Section~\ref{sec:problem}. We then introduce agentic ontology construction in Section~\ref{sec:construction} and describe how the ontology orchestrates the data agent's context and actions in Section~\ref{sec:grounded-action}. Figure~\ref{fig:method-overview} presents an overview of \ours{}.
\subsection{Problem Formulation}
\label{sec:problem}

\paragraph{Data lake and task.}
A data lake $\mathcal{D} = \{f_1, \dots, f_N\}$ is a collection of heterogeneous files. Each file $f$ stores its data under a set of fields $\Phi(f)$, where a field is a named attribute such as a table column or a JSON key. A task is a natural-language question $q$ over $\mathcal{D}$ with a ground-truth answer $a^*$. Answering $q$ requires only a small subset of the lake, which we call the \emph{support} of the task, $S_q \subseteq \mathcal{D}$. The answer is computed by a program $p$ over the support, $a^* = p(S_q)$, which may filter, aggregate, and join data across files in $S_q$ by their fields. Fields, however, are named locally. The same semantic value can appear under different names in different files.

\paragraph{Ontology over the data lake.}
We model the data lake with an ontology $\mathcal{O} = (\mathcal{D}, \Sigma, \mathcal{C}, \mathcal{L})$ that separates how data are stored from what they mean. A schema $\sigma \in \Sigma$ is a set of fields, each with a name, a data type, and a natural-language description of its meaning. A file $f$ instantiates $\sigma$ if every field of $f$ is correctly described by $\sigma$, written as $f \models \sigma$. Files, schemas, and the instantiation relation form the \emph{structural layer}. A concept $c \in \mathcal{C}$ denotes a lake-wide meaning. A link $l = (x, r, c) \in \mathcal{L}$ with relation type $r \in \mathcal{R}$ connects a field of a schema or a concept $x$ to a concept $c$. Concepts and links form the \emph{semantic layer}. Fields of different schemas relate only through the concepts they link to, which makes cross-file alignment explicit.

\subsection{Agentic Ontology Construction}
\label{sec:construction}

\paragraph{Construction through exploration.}
Ontology construction requires recovering domain knowledge that is implicit in the data. Observed values, their distributions, and dependencies between fields provide evidence about the concepts and relationships a data source can support~\citep{gorskis2016database}. A \emph{proposer} explores the lake to interpret these regularities and formulate candidate ontology changes as typed claims. These interpretations remain hypotheses because a pattern in the data does not by itself establish its domain meaning or scope. Verification tests the claims against the underlying sources; accepted claims update the ontology, while contradictory evidence guides further exploration and revision. Construction proceeds from local structure to shared semantics, first inducing schemas and then relating them through concepts. Appendix~\ref{app:algorithms} presents the construction and grounded-action algorithms.

\paragraph{Schema induction.}
The structural stage jointly determines which files share a record structure and how their fields should be interpreted. Given a batch $B \subseteq \mathcal{D}$, the proposer partitions it into disjoint groups $B_1,\dots,B_K$. For each group $B_i$, the proposer induces a schema $\sigma_i$ such that $f \models \sigma_i$ for every $f \in B_i$. The proposer inspects file contents to identify record organization and field types, then interprets the observed values together with available documentation to infer field meanings. Comparing these observations across files establishes the scope over which a common description applies. File grouping and schema description are jointly refined to ensure that each schema accurately represents the structure and meaning of all assigned files. When observations contradict a shared description, the agent revises either the schema or its file assignments. This process consolidates structurally and semantically compatible files into reusable schemas for subsequent semantic analysis.

\paragraph{Semantic enrichment.}
The semantic stage identifies domain meanings that connect the induced schemas. With the structural layer fixed, the proposer explores correspondences among field descriptions and the data they describe, using the schemas to organize its investigation of the underlying sources. This exploration concerns whether locally defined attributes express compatible meanings and what relationship those meanings support. Value correspondences supply evidence for a proposed alignment, while the surrounding schema and documentation constrain its interpretation. The proposer abstracts these interpretations into concepts $c \in \mathcal{C}$ and links $(x,r,c) \in \mathcal{L}$, expressing how fields or concepts relate to a shared meaning. Verification determines which proposed relationships are retained. This stage preserves source-specific descriptions while making their semantic connections explicit, allowing related data to be discovered through concepts across schema boundaries.

\paragraph{Experience distillation.}
Experience distillation retains knowledge acquired during construction that schemas, concepts, and links do not express. After each iteration, a \emph{distiller} summarizes exploration and verification traces into short natural-language lessons stored as \emph{experience} $\mathcal{E}$. Lessons capture data discrepancies and unsuccessful analysis approaches, preserving findings from both accepted and rejected claims for subsequent reasoning. Each lesson $e \in \mathcal{E}$ is anchored to the most specific relevant schemas, fields, or concepts through a set $A_e$. Retrieving an element $y$ during construction or grounding it during task execution (Section~\ref{sec:grounded-action}) supplies the associated lessons $\mathcal{E}(y)=\{e\in\mathcal{E}:y\in A_e\}$. This selective retrieval makes prior findings available when relevant without exhausting the context window.

\subsection{Ontology-Grounded Action}
\label{sec:grounded-action}

During task execution, \ours{} uses the ontology to orchestrate the data agent's observations and actions in the environment. Before processing a raw data file, the agent retrieves its available semantic and structural information from the ontology. Computation is restricted to files in the task graph after the agent interprets the relevant ontology elements and identifies the files as potential support $S_q$ for the task. The agent alternates between ontology observation and task-graph-based action as new observations and intermediate results reveal additional data requirements.

\paragraph{Ontology-based observation.}
The agent observes the data lake through its ontology rather than directly inspecting raw files. Given a task or an intermediate information need, the agent searches for relevant ontology elements and inspects their descriptions to determine whether to add the associated files to the task graph for later computation.
Let $t=0,1,\ldots$ index the agent's execution steps. At step $t$, a query $u_t$ based on the task and observations accumulated so far returns the context
\begin{equation}
\begin{aligned}
H_t &= \mathcal{N}_2\!\left(\operatorname{Search}(\mathcal{O},u_t)\right), \\
K_t &= \left(\operatorname{Describe}(H_t),\;\bigcup_{y\in V(H_t)}\mathcal{E}(y)\right),
\end{aligned}
\label{eq:ontology-context}
\end{equation}
where $\mathcal{N}_2$ forms the subgraph of matched elements and their neighbors within two hops. $\operatorname{Describe}$ returns domain knowledge in $H_t$. Links to neighboring fields explain how this knowledge is represented and organized. Associated experience adds findings from construction that qualify how the sources should be interpreted or used. Based on $K_t$, the agent assesses semantic relevance and whether the available record structure supports the required computation. Repeated search and inspection progressively connect task requirements to candidate sources. 

The agent records the sources selected through observation in a task graph before reading them. The task graph $G_q^t$ records the selected files together with their ontology associations, representing the agent's current hypothesis about the data required for the task, with $G_q^0=(\emptyset,\emptyset)$. Let $\mathcal{I}_t\subseteq\Sigma$ contain the schemas whose descriptions have been inspected. A set of candidate files $F_t$ can be grounded only when
\begin{equation}
\{\sigma\in\Sigma\mid\exists f\in F_t,\ f\models\sigma\}\subseteq\mathcal{I}_t.
\label{eq:schema-before-grounding}
\end{equation}
The agent then selects $F_t$ according to the task requirements and expands the graph through
\begin{equation}
\begin{aligned}
G_q^{t+1} &= G_q^t\cup\operatorname{cl}(F_t), \\
\widehat{S}_q^{t+1} &= R(G_q^{t+1}),
\end{aligned}
\label{eq:support-grounding}
\end{equation}
where $\operatorname{cl}(F_t)$ contains the selected files and their available ontology associations and $R(G)$ denotes the files accessible through $G$. The resulting $\widehat{S}_q^{t+1}$ is the agent's current hypothesis about the support of the task after the grounding update.

\paragraph{Task graph-based action.}
The agent performs computation over the support recorded in its task graph. It produces a program $p_t$ that operates on a subset $S_t$ of the grounded support and returns a result $o_t$,
\begin{equation}
 o_t=p_t(S_t),\qquad S_t\subseteq\widehat{S}_q^t=R(G_q^t).
\label{eq:grounded-access}
\end{equation}
Each $p_t$ contributes to the answer-producing program $p$ in Section~\ref{sec:problem}. Intermediate results can reveal missing data requirements, prompting further ontology observation and grounding before computation continues. This interleaving develops the support hypothesis alongside the computation, while requiring source selection to be informed by schema descriptions. Algorithm~\ref{alg:grounded-action} gives the execution procedure in detail.

\section{Experiments}
\label{sec:experiments}

We evaluate \ours{} on two benchmarks spanning data-to-insight pipelines and scientific hypothesis discovery. We first describe the evaluation setup and then compare its performance with five agent baselines.

\begin{table*}[t]
\centering

\footnotesize
\setlength{\tabcolsep}{2pt}
\renewcommand{\arraystretch}{1.14}
\begin{tabular*}{\textwidth}{@{\extracolsep{\fill}}cl*{5}{rr}@{}}
\toprule
\multirow{3}{*}{\textbf{Model}} & \multirow{3}{*}{\textbf{Method}} & \multicolumn{4}{c}{KramaBench} & \multicolumn{6}{c}{DiscoveryBench} \\
\cmidrule(lr){3-6}\cmidrule(lr){7-12}
 & & \multicolumn{2}{c}{Answer} & \multicolumn{2}{c}{Pipeline} & \multicolumn{2}{c}{HMS} & \multicolumn{2}{c}{ctxF1} & \multicolumn{2}{c}{varF1} \\
\cmidrule(lr){3-4}\cmidrule(lr){5-6}\cmidrule(lr){7-8}\cmidrule(lr){9-10}\cmidrule(lr){11-12}
 & & Score$\uparrow$ & $\Delta\uparrow$ & Score$\uparrow$ & $\Delta\uparrow$ & Score$\uparrow$ & $\Delta\uparrow$ & Score$\uparrow$ & $\Delta\uparrow$ & Score$\uparrow$ & $\Delta\uparrow$ \\
\midrule
\multirow{6}{*}{\shortstack{DeepSeek-\\V4.1-Flash}} & Smolagents & $56.40$ & $-$ & $43.12$ & $-$ & $11.72$ & $-$ & $\underline{28.87}$ & $-$ & $68.09$ & $-$ \\
 & Reflexion & $\underline{62.18}$ & $\underline{+5.78}$ & $\underline{52.73}$ & $\underline{+9.61}$ & $9.62$ & $-2.10$ & $25.37$ & $-3.50$ & $68.22$ & $+0.13$ \\
 & DS-STAR & $43.66$ & $-12.74$ & $38.42$ & $-4.70$ & $11.46$ & $-0.26$ & $22.69$ & $-6.18$ & $\underline{76.10}$ & $\underline{+8.01}$ \\
 & Data Interpreter & $56.16$ & $-0.24$ & $47.91$ & $+4.79$ & $\underline{12.58}$ & $\underline{+0.86}$ & $26.08$ & $-2.79$ & $75.92$ & $+7.83$ \\
 & DatawiseAgent & $56.83$ & $+0.43$ & $\mathbf{52.86}$ & $\mathbf{+9.74}$ & $11.57$ & $-0.15$ & $25.53$ & $-3.34$ & $74.28$ & $+6.19$ \\
\rowcolor{oursblue}
 & \ours{} (ours) & $\mathbf{73.00}$ & $\mathbf{+16.60}$ & $49.77$ & $+6.65$ & $\mathbf{17.07}$ & $\mathbf{+5.35}$ & $\mathbf{31.65}$ & $\mathbf{+2.78}$ & $\mathbf{79.55}$ & $\mathbf{+11.46}$ \\
\midrule
\multirow{6}{*}{GPT-6-Luna} & Smolagents & $44.13$ & $-$ & $38.97$ & $-$ & $8.81$ & $-$ & $\underline{24.15}$ & $-$ & $77.94$ & $-$ \\
 & Reflexion & $40.69$ & $-3.44$ & $43.78$ & $+4.81$ & $9.80$ & $+0.99$ & $21.89$ & $-2.26$ & $78.91$ & $+0.97$ \\
 & DS-STAR & $48.42$ & $+4.29$ & $38.69$ & $-0.28$ & $\underline{10.85}$ & $\underline{+2.04}$ & $24.08$ & $-0.07$ & $\underline{83.52}$ & $\underline{+5.58}$ \\
 & Data Interpreter & $45.93$ & $+1.80$ & $\underline{46.50}$ & $\underline{+7.53}$ & $8.81$ & $+0.00$ & $20.64$ & $-3.51$ & $77.92$ & $-0.02$ \\
 & DatawiseAgent & $\underline{49.66}$ & $\underline{+5.53}$ & $\mathbf{48.30}$ & $\mathbf{+9.33}$ & $9.59$ & $+0.78$ & $23.10$ & $-1.05$ & $79.36$ & $+1.42$ \\
\rowcolor{oursblue}
 & \ours{} (ours) & $\mathbf{52.96}$ & $\mathbf{+8.83}$ & $46.39$ & $+7.42$ & $\mathbf{13.35}$ & $\mathbf{+4.54}$ & $\mathbf{26.57}$ & $\mathbf{+2.42}$ & $\mathbf{83.94}$ & $\mathbf{+6.00}$ \\
\midrule
\multirow{6}{*}{MiMo-V2.6-Pro} & Smolagents & $57.83$ & $-$ & $\mathbf{49.38}$ & $-$ & $10.92$ & $-$ & $29.80$ & $-$ & $65.42$ & $-$ \\
 & Reflexion & $56.74$ & $-1.09$ & $45.88$ & $-3.50$ & $6.18$ & $-4.74$ & $19.83$ & $-9.97$ & $61.98$ & $-3.44$ \\
 & DS-STAR & $\underline{58.08}$ & $\underline{+0.25}$ & $41.43$ & $-7.95$ & $11.72$ & $+0.80$ & $27.76$ & $-2.04$ & $\underline{77.16}$ & $\underline{+11.74}$ \\
 & Data Interpreter & $55.31$ & $-2.52$ & $\underline{48.66}$ & $\underline{-0.72}$ & $11.89$ & $+0.97$ & $29.92$ & $+0.12$ & $72.56$ & $+7.14$ \\
 & DatawiseAgent & $52.22$ & $-5.61$ & $45.54$ & $-3.84$ & $\underline{13.31}$ & $\underline{+2.39}$ & $\underline{30.54}$ & $\underline{+0.74}$ & $72.70$ & $+7.28$ \\
\rowcolor{oursblue}
 & \ours{} (ours) & $\mathbf{68.54}$ & $\mathbf{+10.71}$ & $48.53$ & $-0.85$ & $\mathbf{17.23}$ & $\mathbf{+6.31}$ & $\mathbf{33.77}$ & $\mathbf{+3.97}$ & $\mathbf{79.45}$ & $\mathbf{+14.03}$ \\
\bottomrule
\end{tabular*}
\caption{\textbf{Performance across two benchmarks and three model backbones.} For each metric, $\Delta$ is the percentage-point difference from smolagents with the same backbone; a dash denotes the reference. Bold and underlining mark the highest and second-highest scores for each backbone--metric pair. Light blue highlights \ours{}.}
\label{tab:overall}
\end{table*}
\subsection{Experimental Setup}
\label{sec:setup}

\paragraph{Benchmarks.}
We evaluate \ours{} on two benchmarks for full-pipeline data analysis over real-world data. KramaBench~\citep{lai2026kramabench} contains 104 curated data-to-insight tasks over about 1{,}700 real-world files from 24 sources in six domains, requiring file discovery, cleaning, and integration before analysis. DiscoveryBench~\citep{majumder2025discoverybench} evaluates data-driven hypothesis discovery. We use its 239-task real-data test split, drawn from published studies.
\paragraph{Metrics.}
For KramaBench, we report the mean answer score using the designated metric for each answer type and the pipeline score measuring coverage of required functionalities~\citep{lai2026kramabench}.
For DiscoveryBench, we report Hypothesis Matching Score (HMS), context F1 (ctxF1), and variable F1 (varF1), which assess overall hypothesis agreement, context matching, and variable matching, respectively~\citep{majumder2025discoverybench}.
We use DeepSeek-V4.1-Flash as the LLM evaluator for all methods. Appendix~\ref{app:metric-implementation} provides scoring details.

\paragraph{Baselines.}
We compare against five agent baselines that differ in how they plan and revise data-analysis workflows.
\textbf{Smolagents}~\citep{smolagents} is a ReAct-style~\citep{yao2023react} coding agent that interleaves reasoning, code execution, and observations.
\textbf{Reflexion}~\citep{shinn2023reflexion} uses verbal self-reflection on task feedback to improve subsequent attempts.
\textbf{Data Interpreter}~\citep{hong-etal-2025-data} decomposes a task into a hierarchical graph of subtasks and executes them with code.
\textbf{DS-STAR}~\citep{nam2026dsstar} iteratively plans, executes, and verifies data-science steps over heterogeneous files.
\textbf{DatawiseAgent}~\citep{you-etal-2025-datawiseagent} provides a notebook-centric framework for data-science automation.
We evaluate all methods with DeepSeek-V4.1-Flash~\citep{xu2026deepseek}, GPT-6-Luna~\citep{openai2026gpt6luna}, and MiMo-V2.6-Pro~\citep{xiaomi2026mimo}. For the main results in Table~\ref{tab:overall}, each \ours{} backbone constructs its own ontology and serves all roles during ontology construction and task execution. The cross-backbone reuse experiment in Section~\ref{sec:ontology-transfer} instead uses an ontology constructed by DeepSeek-V4.1-Flash.

\paragraph{Implementation.}
For \ours{}, we allow the agent a maximum of 30 steps during task execution and 20 steps during ontology construction stages. We implement Reflexion on top of \texttt{smolagents}, augmenting its ReAct-style coding agent with verbal self-reflection between attempts. We allow a maximum of 30 agent steps for both smolagents and Reflexion. DS-STAR is limited to 10 steps and four revisions, while Data Interpreter may revise its plan up to four times. All methods execute code in the same isolated sandbox, with access to the data lake and no internet access. Appendix~\ref{app:baseline-details} specifies generation and baseline settings, and Appendix~\ref{app:ours-implementation} provides further implementation details for \ours{}.

\subsection{Main Results}
\label{sec:main-results}
\textbf{\ours{} consistently outperforms the strongest baselines across backbones and benchmarks.}
Table~\ref{tab:overall} shows that \ours{} achieves the highest mean KramaBench answer score and DiscoveryBench HMS for every evaluated backbone. Relative to the strongest baseline for each backbone, \ours{} improves KramaBench answer scores by 10.82, 3.30, and 10.46 percentage points with DeepSeek-V4.1-Flash, GPT-6-Luna, and MiMo-V2.6-Pro, respectively. The corresponding gains in DiscoveryBench HMS are 4.49, 2.50, and 3.92 percentage points. \ours{} consistently maintains its advantage, suggesting that ontology grounding benefits backbones of differing capability rather than depending on a particular one. While the gains in KramaBench pipeline score are less consistent, the improvements in final answer quality across both benchmarks suggest that explicitly modeling data semantics and grounding action in relevant sources provides a robust foundation for heterogeneous data analysis.

\section{Analysis}
\label{sec:analysis}

We examine three research questions about \ours{}: whether agent-constructed ontologies guide exploration to the gold-standard files required by a task, whether a constructed ontology can be reused across generation backbones, and which components contribute to answer quality.

\subsection{RQ1: Does the ontology guide exploration to the task support files?}
\label{sec:ontology-quality}

We examine whether the constructed ontology guides exploration toward the gold-standard files needed to answer a task, i.e., the task's support $S_q$.
Table~\ref{tab:file-read-references} in Appendix~\ref{app:file-read-references} directly evaluates this aspect through explicit file-reading references in saved code across 104 tasks and 311 required input units.
 
\ours{} achieves the highest explicit-unit coverage with GPT-6-Luna and MiMo-V2.6-Pro and approaches the strongest baseline with DeepSeek-V4.1-Flash. With GPT-6-Luna, it covers 143/311 units (45.98\%), exceeding the strongest baseline, DatawiseAgent (43.73\%), by 2.25 percentage points. With MiMo-V2.6-Pro, it covers 179/311 units (57.56\%), exceeding DatawiseAgent (49.52\%) by 8.04 points.
With DeepSeek-V4.1-Flash, it also covers 182/311 units (58.52\%), ranking second to DatawiseAgent (60.45\%) with a gap of 1.93 percentage points. These results indicate that ontology grounding steers the agent toward the task’s support files more consistently than the direct exploration used by the baselines.

To examine whether our ontology construction paradigm contributes to hypotheses in the correct context, we compare smolagents with \ours{} using either a rule-based ontology or one constructed with DeepSeek-V4.1-Flash on DiscoveryBench, shown in Figure~\ref{fig:task-scatter}. The rule-based construction procedure in Appendix~\ref{app:ontology} provides a deterministic alternative to agentic construction. The vertical axis measures variable and relationship correctness within matched hypothesis pairs. With the constructed ontology, \ours{} moves tasks toward the upper-right corner of Figure~\ref{fig:task-scatter}, where both context F1 and within-pair correctness reach 100\%. The rule-based ontology achieves an HMS of 11.63\%, compared with 11.72\% for smolagents and 17.07\% for the agentically constructed ontology. Grounding in an ontology that records only file headers thus provides little benefit. The gain depends on the field meanings and cross-file relations that agentic construction adds, so grounding and construction quality are complementary rather than interchangeable.

\begin{figure}[tbp]
    \centering
    \includegraphics[width=\linewidth]{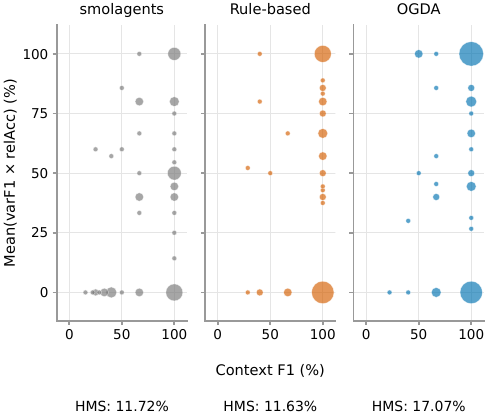}
    \caption{Task-level hypothesis quality on DiscoveryBench. From left to right: smolagents, \ours{} with a rule-based ontology, and \ours{} with an ontology built by DeepSeek-V4.1-Flash. Bubble area is proportional to the number of tasks sharing a coordinate.}
    \label{fig:task-scatter}
\end{figure}
 
\subsection{RQ2: Can an ontology be reused across backbones?}
\label{sec:ontology-transfer}
We test whether an ontology constructed by one model remains useful when another model performs the analysis.  We construct the ontology once with DeepSeek-V4.1-Flash and reuse it with four generation backbones on KramaBench. Details are in Appendix~\ref{app:ontology-transfer}.  We compare against smolagents, which uses no ontology, and EvoOntology~\citep{chong2026evoontologyselfevolvingontologylayer}, which builds an ontology layer for data agents from sampled agent traces.

\begin{figure}[tbp]
    \centering 
    \includegraphics[width=\linewidth]{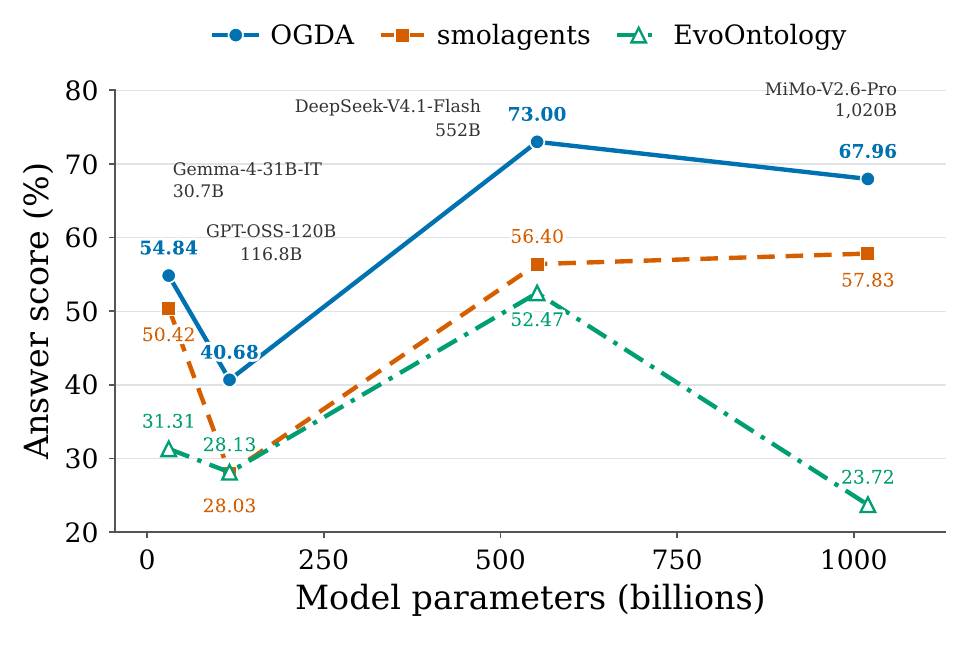}
    \caption{Ontology reuse across generation backbones on KramaBench. We compare \ours{}, smolagents, and EvoOntology with Gemma-4-31B-IT, GPT-OSS-120B, DeepSeek-V4.1-Flash, and MiMo-V2.6-Pro.}
    \label{fig:answer-scores-backbones}
\end{figure}

As shown in Figure~\ref{fig:answer-scores-backbones}, \ours{} achieves the highest answer score with every backbone. It improves over smolagents by 4.42, 12.65, 16.60, and 10.13 percentage points with Gemma-4-31B-IT~\citep{gemmateam2026gemma4}, GPT-OSS-120B~\citep{openai2025gptoss120bgptoss20bmodel}, DeepSeek-V4.1-Flash, and MiMo-V2.6-Pro. A single construction can therefore be shared across backbones rather than repeated for each.
  
\subsection{RQ3: Which components improve answer quality? Ablation Study}
\label{sec:ablation}
We ablate the four components of \ours{}, the ontology-grounded agent, verifier, semantic layer, and experience, to examine their contributions to KramaBench answer quality. Experimental details are provided in Appendix~\ref{app:ablation}.

\paragraph{Ontology-Grounded Agent.}
Replacing the ontology-grounded agent with smolagents produces the largest observed decrease, from 73.00\% to 60.94\%. This variant retains read-only access to the ontology, and it still outperforms plain smolagents by 4.54 points. However, 12.06 of \ours{}'s 16.60-point gain over smolagents appears only when the ontology also governs data access.  Without the task graph's constraints, ontology information is merely advisory. More specifically, the agent can read files whose schemas it has not inspected, and source-selection errors propagate into the computation.

\paragraph{Verifier.}
Removing the verifier during ontology construction reduces the answer score to 66.61\%, a decrease of 6.39 percentage points compared with the full system. In \ours{}, the verifier independently tests claims inferred from file contents against the underlying data. The decrease suggests that independent verification during construction helps produce an ontology that reliably supports downstream analysis.
\begin{figure}[tbp]
    \centering
    \includegraphics[width=\linewidth,trim=35 40 15 15,clip]{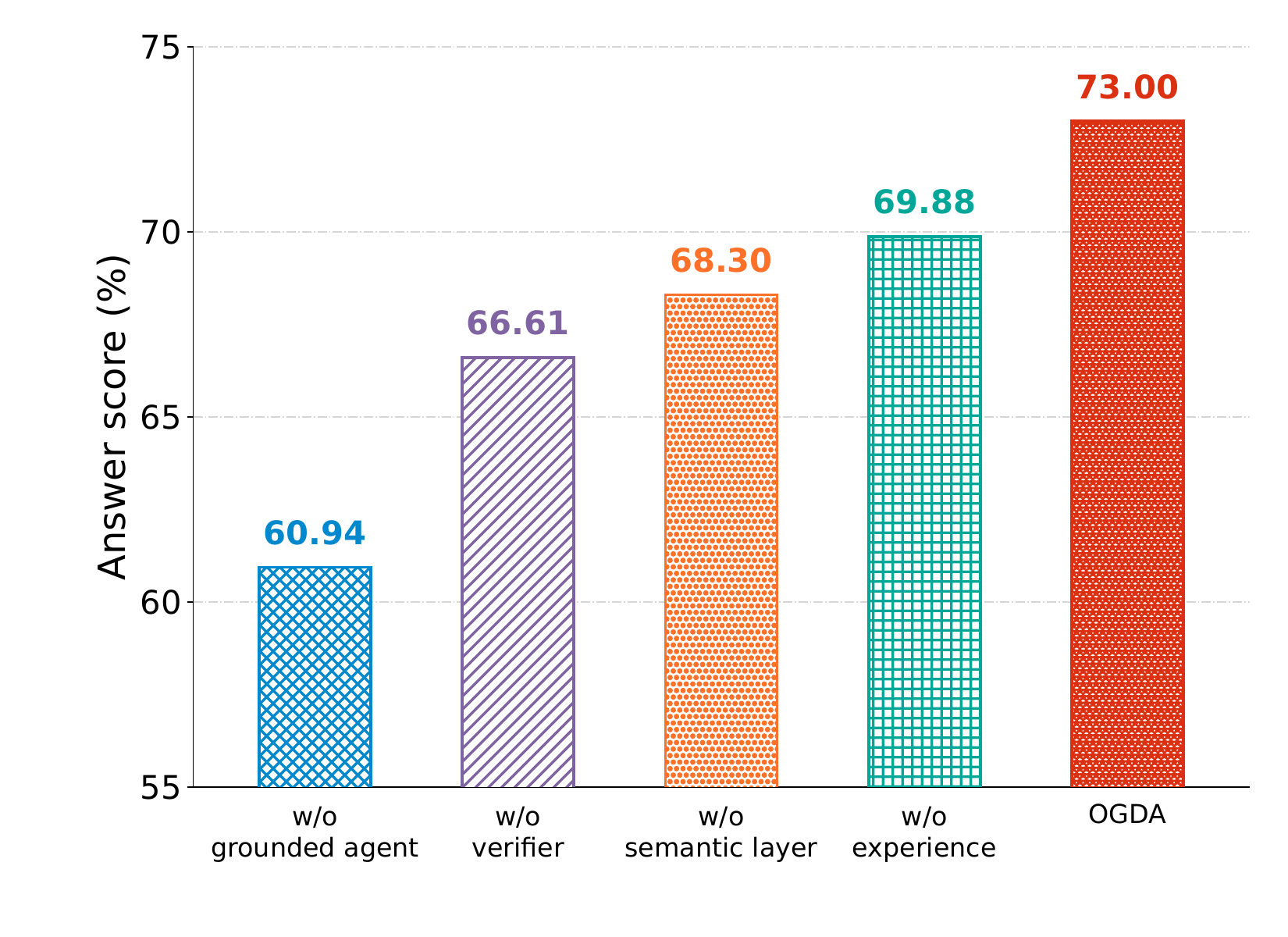}
    \caption{Component ablations on KramaBench with DeepSeek-V4.1-Flash. Each variant removes one component from \ours{}. Values are answer scores (\%).}
    \label{fig:ablation-accuracy}
\end{figure}
\paragraph{Semantic layer.}
Removing the semantic layer, retaining schemas but dropping the lake-wide concepts and the links that connect them, reduces the answer score to 68.30\%. The observed decrease suggests that representing these meanings is useful beyond organizing local structure alone.

\paragraph{Experience.}
Removing the distiller eliminates experience from both ontology construction and task execution, reducing the answer score to 69.88\%. Experience records knowledge that the ontology cannot express as claims, such as discrepancies between documentation and data and analysis approaches that failed. The decrease suggests that producing and reusing these lessons across construction and execution provides useful context beyond the schemas, concepts, and relations explicitly represented in the ontology.

\section{Conclusion}
\label{sec:conclusion}
We introduced \ours{}, an ontology-grounded framework for data analysis over heterogeneous data lakes. \ours{} builds an ontology through a hypothesize-and-verify loop and grounds each analysis in a task graph that makes source selection explicit and governs data access. Across KramaBench and DiscoveryBench with three LLM backbones, \ours{} achieves the highest answer scores and HMS among the evaluated methods, and ablations identify grounded execution as the largest contributor.

\section*{Limitations}
OGDA is designed for lakes of record-structured data, such as tables, spreadsheets, databases, and semi-structured files, with free text serving mainly as evidence for field meanings. It does not model non-textual modalities such as images, audio, or video. These sources lack the fields and values that schema induction and concept alignment rely on. Extending the ontology to multimodal sources would require new schema abstractions and verification procedures for content that cannot be checked through value correspondences. In addition, OGDA's answers can only be as reliable as the ontology and the data beneath it. Verification tests proposed claims against the lake itself, so noisy, inconsistent, or incomplete data can yield ontology elements that are consistent with the observed values but do not reflect their intended meaning. The same noise affects computation at execution time. Grounding narrows the agent's attention to the right sources, but it cannot correct errors within those sources, so OGDA does not guarantee correct outputs when the underlying data are unreliable.
\bibliography{anthology-1,anthology-2,custom}
\appendix
\label{sec:appendix}
\section{Algorithms for \ours{}}
\label{app:algorithms}

Algorithms~\ref{alg:construction} and~\ref{alg:grounded-action} give the procedures of Sections~\ref{sec:construction} and~\ref{sec:grounded-action}.
Algorithm~\ref{alg:construction} observes only the data lake $\mathcal{D}$ and never a task; its first stage induces the structural layer and its second stage enriches the semantic layer.
Algorithm~\ref{alg:grounded-action} answers a task with the resulting ontology and experience; the agent reads only files placed in its task graph.
\begin{algorithm}[t]
\caption{Agentic ontology construction}
\label{alg:construction}
\begin{algorithmic}[1]
\Require Data lake $\mathcal{D}$ (task-blind), proposer $\mathcal{P}$, verifier $\mathcal{V}$, distiller $\mathcal{X}$, maximum passes $T$
\Ensure Ontology $\mathcal{O}$ and experience $\mathcal{E}$
\State $\mathcal{O} \gets \emptyset$; \quad $\mathcal{E} \gets \emptyset$
\Statex \textit{Stage 1: structural induction}
\ForAll{batches $B \subseteq \mathcal{D}$}
    \State $H \gets \mathcal{P}(B, \mathcal{O})$ \Comment{cluster files; one schema per cluster}
    \State revise $H$ until it passes executable checks on $\mathcal{D}$
    \State $\mathcal{O} \gets \mathcal{O} \oplus \{\, h \in H : h \text{ accepted} \,\}$ \Comment{documented, or judged by $\mathcal{V}$}
    \State $\mathcal{E} \gets \mathcal{E} \cup \mathcal{X}(H)$
\EndFor
\Statex \textit{Stage 2: semantic enrichment}
\For{$t = 1, \dots, T$}
    \State $H \gets \mathcal{P}.\textsc{Enrich}(\mathcal{O}, \mathcal{E})$ \Comment{concepts, links, removals}
    \State test and revise $H$ with $\mathcal{V}$, including Eq.~\eqref{eq:overlap} for key-field links
    \State $H_{\mathrm{acc}} \gets \{\, h \in H : h \text{ accepted} \,\}$
    \State $\mathcal{O} \gets \mathcal{O} \oplus H_{\mathrm{acc}}$; \quad $\mathcal{E} \gets \mathcal{E} \cup \mathcal{X}(H)$
    \If{$H_{\mathrm{acc}} = \emptyset$} \textbf{break} \EndIf
\EndFor
\State \Return $\mathcal{O}, \mathcal{E}$
\end{algorithmic}
\end{algorithm}

For the key-field overlap test in Algorithm~\ref{alg:construction}, $\mathrm{val}(\phi)$ denotes the set of distinct values of field $\phi$ across files instantiating its schema, and $\theta$ is the overlap threshold:
\begin{equation}
\mathrm{ov}(\phi, \phi') = \frac{|\mathrm{val}(\phi) \cap \mathrm{val}(\phi')|}{\min\big(|\mathrm{val}(\phi)|, |\mathrm{val}(\phi')|\big)} \;\ge\; \theta .
\label{eq:overlap}
\end{equation}
\begin{algorithm}[t]
\caption{Ontology-grounded action}
\label{alg:grounded-action}
\begin{algorithmic}[1]
\Require Task $q$, ontology $\mathcal{O}$, experience $\mathcal{E}$, lake $\mathcal{D}$, agent $\pi$, step budget $M$
\Ensure Answer $\hat{a}$ and task graph $G_q$
\State $G_q \gets (\emptyset, \emptyset)$; \quad $\mathcal{I} \gets \emptyset$; \quad $\tau \gets (q)$
\For{$n = 1, \dots, M$}
    \State $p_n \gets \pi(\tau)$ \Comment{search, inspect, ground, read, write, or answer}
    \If{$p_n$ inspects schema $\sigma$}
        \State $o_n \gets$ schema description and associated experience
        \State $\mathcal{I} \gets \mathcal{I} \cup \{\sigma\}$
    \ElsIf{$p_n$ grounds files $F$}
        \State $\Sigma_F \gets \{\sigma\in\Sigma\mid\exists f\in F,\ f\models\sigma\}$
        \If{$\Sigma_F\nsubseteq\mathcal{I}$}
            \State $o_n \gets$ ``inspect the required schema descriptions first''
        \Else
            \State $G_q' \gets G_q \cup \mathrm{cl}(F)$
            \State $o_n \gets$ new elements, files, and lessons $\mathcal{E}(G_q') \setminus \mathcal{E}(G_q)$
            \State $G_q \gets G_q'$
        \EndIf
    \ElsIf{$p_n$ reads a file $f \notin R(G_q)$}
        \State $o_n \gets$ ``ground $f$ first''
    \ElsIf{$p_n$ answers $\hat{a}$}
        \State add $\hat{a}$ to $G_q$, derived from the data read; \Return $\hat{a}, G_q$
    \Else
        \State $o_n \gets \textsc{Execute}(p_n)$ \Comment{results enter $G_q$ as derived nodes}
    \EndIf
    \State $\tau \gets \tau \cdot (p_n, o_n)$
\EndFor
\State \Return \textsc{None}, $G_q$
\end{algorithmic}
\end{algorithm}

\section{Experiment Details}
\label{app:experiment-details}

\subsection{Metric Implementation}
\label{app:metric-implementation}

\paragraph{Evaluator configuration.}
We use DeepSeek-V4.1-Flash with high reasoning effort and a maximum output length of 16k tokens per evaluator call for all LLM-based judgments on KramaBench and DiscoveryBench.
This evaluator is shared across all methods and generation backbones.
We express the resulting benchmark scores as percentages.

\paragraph{KramaBench answer score.}
Each task receives a score $s_i\in[0,1]$ using the benchmark's designated primary metric for its answer type~\citep{lai2026kramabench}.
Exact string and numeric answers use binary correctness, approximate numeric answers use a relative-error score, and list answers use F1 with the prescribed item-matching criterion.
The headline answer score gives every task equal weight:
\begin{equation}
S_{\mathrm{answer}} = \frac{100}{104}\sum_{i=1}^{104}s_i.
\label{eq:krama-answer-score}
\end{equation}
Of these 104 tasks, 10 (approximately 9.6\%) require an LLM judge, while the remaining 94 (approximately 90.4\%) are graded deterministically.
For each of the 10 LLM-graded tasks, the judge answers a binary yes/no question about answer correctness, mapped to a score of 1 or 0.

\paragraph{KramaBench pipeline score.}
Pipeline evaluation separately checks the generated code against the human-annotated key functionalities for each task~\citep{lai2026kramabench}.
For task $i$, let $K_i$ denote this set and let $b_{ik}\in\{0,1\}$ indicate the judge's decision on whether the code covers functionality $k$.
The task-level pipeline score is
\begin{equation}
p_i = \frac{1}{|K_i|}\sum_{k\in K_i}b_{ik}.
\end{equation}
We average these task-level scores and multiply by 100.
Thus, tasks receive equal weight regardless of how many functionalities they contain.
The 10/94 split applies only to answer evaluation; pipeline evaluation uses LLM judgments of functionality coverage.

\paragraph{DiscoveryBench metrics.}
We evaluate generated hypotheses against gold hypotheses using context F1 (ctxF1), variable F1 (varF1), and Hypothesis Matching Score (HMS)~\citep{majumder2025discoverybench}.
The evaluator decomposes each hypothesis into sub-hypotheses with a context, interacting variables, and a relationship. If extraction returns an empty set, it uses the original hypothesis as a fallback sub-hypothesis.
It then forms one-to-one matches between generated and gold sub-hypotheses whose contexts are semantically equivalent.
Let $n_p$ and $n_g$ be the numbers of generated and gold sub-hypotheses, and let $\mathcal{M}$ be the matched pairs.
For nonempty generated and gold sets, context precision and recall are $P_c=|\mathcal{M}|/n_p$ and $R_c=|\mathcal{M}|/n_g$, giving
\begin{equation}
\mathrm{ctxF1}=\frac{2P_cR_c}{P_c+R_c}.
\end{equation}
Within each matched pair $m$, the judge identifies the variables in both sub-hypotheses and their semantic overlap, accounting for paraphrases.
Their precision and recall yield a variable F1 score $v_m$.
The relationship score $r_m$ is 1 for an equivalent relationship, 0.5 for a similar but less specific relationship, and 0 for a different relationship.
For tasks with at least one matched pair, the conditional variable F1 and task-level HMS are
\begin{equation}
\begin{aligned}
\mathrm{varF1} &= \frac{1}{|\mathcal{M}|}\sum_{m\in\mathcal{M}}v_m,\\
\mathrm{HMS} &= R_c\cdot\frac{1}{n_p}
\sum_{m\in\mathcal{M}}v_mr_m.
\end{aligned}
\end{equation}
HMS multiplies context recall by the accuracy averaged over all generated sub-hypotheses; unmatched generated sub-hypotheses contribute zero. When no contexts match, ctxF1 and HMS are zero, while the within-pair variable F1 is undefined. Thus, HMS rewards correct variables and relationships while penalizing both missing gold contexts and unmatched generated contexts. Benchmark HMS is the mean task-level HMS over all 239 tasks, multiplied by 100.

\subsection{Baselines}
\label{app:baseline-details}

For the main experiments, all three generation backbones---DeepSeek-V4.1-Flash, GPT-6-Luna, and MiMo-V2.6-Pro---run with thinking disabled and otherwise use their default generation parameters across all methods.
We use the \texttt{CodeAgent} variant of smolagents, which permits and encourages combining multiple logical operations in a single code block within one agent step.
Our Reflexion implementation builds on the same \texttt{CodeAgent} and follows the same convention, adding verbal self-reflection between attempts.

\subsection{Implementation Details of \ours{}}
\label{app:ours-implementation}

For each construction backbone, we model each data lake once and reuse the resulting ontology and experience across all tasks on that lake.
The proposer and verifier each have a maximum budget of 20 agent steps per invocation.
We use batches of 8 files for schema induction and 32 claims for verification.
Ontology search ranks elements by cosine similarity between bag-of-words representations of the query and ontology elements, retrieves the top 10 matches, and expands each match to include its two-hop neighborhood and connecting relations.

\section{Ontology Details}
\label{app:ontology}

\paragraph{Constructed ontology statistics.}
Table~\ref{tab:ontology-statistics} reports the numbers of schemas, fields, tables, concepts, and links constructed by each model, aggregated over six KramaBench lakes and ten DiscoveryBench lakes.

\begin{table*}[t]
\centering
\small
\setlength{\tabcolsep}{5pt}
\renewcommand{\arraystretch}{1.15}
\begin{tabularx}{\textwidth}{@{}lXrrrrrr@{}}
\toprule
\textbf{Benchmark} & \textbf{Construction model} & \textbf{Lakes} & \textbf{Schemas} & \textbf{Fields} & \textbf{Tables} & \textbf{Concepts} & \textbf{Links} \\
\midrule
\multirow{3}{*}{KramaBench} & DeepSeek-V4.1-Flash & 6 & 100 & 1,577 & 1,723 & 294 & 1,791 \\
 & GPT-6-Luna & 6 & 78 & 1,487 & 602 & 33 & 136 \\
 & MiMo-V2.6-Pro & 6 & 80 & 779 & 1,706 & 116 & 336 \\
\midrule
\multirow{3}{*}{DiscoveryBench} & DeepSeek-V4.1-Flash & 10 & 21 & 837 & 21 & 210 & 1,285 \\
 & GPT-6-Luna & 10 & 20 & 800 & 20 & 88 & 265 \\
 & MiMo-V2.6-Pro & 10 & 21 & 838 & 21 & 183 & 554 \\
\bottomrule
\end{tabularx}
\caption{Constructed ontology statistics by benchmark and construction model. Counts are totals across the listed lakes.}
\label{tab:ontology-statistics}
\end{table*}

\paragraph{Rule-based ontology.}
\label{app:rule-based-ontology}
To measure what agentic construction contributes beyond the structure that a deterministic profiler can recover, we replace the construction loop of Section~\ref{sec:construction} with a fixed set of rules that involve no model calls.
Let $\mathcal{D}_{\mathrm{tab}} \subseteq \mathcal{D}$ denote the delimited-text and spreadsheet files of the lake, where each sheet of a workbook is treated as a separate file.
For a file $f$, let $\mathrm{hdr}(f) = (u_1, \dots, u_m)$ be the cells of its first row.
The rules assume that this row is the header and admit $f$ only if $\mathrm{hdr}(f)$ contains at least two non-empty cells.
This is the rule-based variant evaluated in Figure~\ref{fig:task-scatter}. Each admitted file receives its own schema $\sigma_f$, with $f \models \sigma_f$:
\begin{equation}
\begin{aligned}
\Sigma_{\mathrm{rule}} = \{\, \sigma_f :{}& f \in \mathcal{D}_{\mathrm{tab}},\\
                         & |\{ j : u_j \neq \varepsilon \}| \geq 2 \,\}.
\end{aligned}
\end{equation}
Here, $\varepsilon$ denotes an empty cell.
The fields of $\sigma_f$ are the distinct non-empty header cells of $f$, in order of first occurrence.
A field is typed as numeric if every non-empty value among the first 20 records of its column parses as a number, and as a string otherwise.
Schemas and fields carry no descriptions, no documentation is associated with any schema, and the semantic layer and the experience store are empty, $\mathcal{C} = \mathcal{L} = \mathcal{E} = \emptyset$.
Files that are not admitted, such as documentation, semi-structured records, and free text, remain outside the ontology; the data agent can still ground them as raw files (Section~\ref{sec:grounded-action}).

The rule-based ontology thus omits several decisions made during agentic construction: clustering files that hold the same kind of record into one schema, locating headers and record regions that do not start at the first row, attaching documentation and deriving field meanings from it, building the semantic layer, and distilling experience.
\section{File-Reading References in Saved Code}
\label{app:file-read-references}
\begin{table*}[t]
\centering
\scriptsize
\setlength{\tabcolsep}{3pt}
\renewcommand{\arraystretch}{1.15}
\begin{tabularx}{\textwidth}{@{}lXrrrrrrrrr@{}}
\toprule
\textbf{Model} & \textbf{Method} & \shortstack{\textbf{Saved}\\\textbf{answers}} & \textbf{Full} & \textbf{Partial} & \shortstack{\textbf{Dynamic}\\\textbf{only}} & \shortstack{\textbf{No}\\\textbf{evidence}} & \shortstack{\textbf{Empty}\\\textbf{code}} & \shortstack{\textbf{Missing}\\\textbf{output}} & \shortstack{\textbf{Explicit}\\\textbf{units / 311}} & \shortstack{\textbf{Explicit}\\\textbf{\%}} \\
\midrule
\multirow{6}{*}{\shortstack[l]{DeepSeek-\\V4.1-Flash}} & smolagents & 90/104 & 49 & 22 & 7 & 12 & 14 & 0 & 106/311 & 34.08 \\
 & Reflexion & 104/104 & 59 & 19 & 6 & 20 & 0 & 0 & 134/311 & 43.09 \\
 & DS-STAR & 91/104 & 55 & 20 & 7 & 10 & 12 & 0 & 115/311 & 36.98 \\
 & Data Interpreter & 102/104 & 60 & 26 & 10 & 8 & 0 & 0 & 141/311 & 45.34 \\
 & \ours{} & 103/104 & 67 & 25 & 5 & 7 & 0 & 0 & 182/311 & 58.52 \\
 & DatawiseAgent & 89/104 & 70 & 26 & 3 & 5 & 0 & 0 & 188/311 & 60.45 \\
\midrule
\multirow{6}{*}{GPT-6-Luna} & smolagents & 104/104 & 38 & 17 & 7 & 42 & 0 & 0 & 98/311 & 31.51 \\
 & Reflexion & 104/104 & 47 & 20 & 8 & 29 & 0 & 0 & 108/311 & 34.73 \\
 & DS-STAR & 86/104 & 44 & 19 & 9 & 20 & 12 & 0 & 87/311 & 27.97 \\
 & Data Interpreter & 95/104 & 56 & 27 & 6 & 15 & 0 & 0 & 126/311 & 40.51 \\
 & \ours{} & 104/104 & 50 & 28 & 4 & 22 & 0 & 0 & 143/311 & 45.98 \\
 & DatawiseAgent & 99/104 & 62 & 24 & 6 & 10 & 2 & 0 & 136/311 & 43.73 \\
\midrule
\multirow{6}{*}{MiMo-V2.6-Pro} & smolagents & 101/104 & 51 & 21 & 5 & 24 & 2 & 1 & 120/311 & 38.59 \\
 & Reflexion & 97/104 & 55 & 23 & 14 & 8 & 4 & 0 & 146/311 & 46.95 \\
 & DS-STAR & 92/104 & 53 & 21 & 5 & 13 & 12 & 0 & 142/311 & 45.66 \\
 & Data Interpreter & 98/104 & 64 & 25 & 7 & 8 & 0 & 0 & 153/311 & 49.20 \\
 & \ours{} & 103/104 & 65 & 25 & 4 & 10 & 0 & 0 & 179/311 & 57.56 \\
 & DatawiseAgent & 82/104 & 57 & 24 & 2 & 5 & 16 & 0 & 154/311 & 49.52 \\
\bottomrule
\end{tabularx}
\caption{File-reading references in saved code. Task categories are counts over 104 tasks; explicit-unit coverage is reported over 311 required units. Saved answers are reported separately and do not establish successful execution or file use. Only the supplied model--method results are shown.}
\label{tab:file-read-references}
\end{table*}
Table~\ref{tab:file-read-references} reports file-reading references identified in saved code across 104 tasks and 311 required input units.
These percentages measure references in saved code, not confirmed file use during successful execution.
A read reference may be exploratory, conditional, unexecuted, or part of an unsuccessful attempt; it does not establish that the file's contents contributed to the final answer.
Comments, filenames mentioned in search, grounding, or listing calls, and answer text are excluded as read evidence.
The analysis traces simple variable paths and literal code nested in \texttt{exec} or \texttt{compile}, but arbitrary runtime values, branch assignments, imported helpers, and complex transformations may remain unresolved.

\emph{Full} means that every required unit has an explicit filename in a traced read call; \emph{Partial} means that some do.
\emph{Dynamic only} denotes family or pattern reads without explicit filenames.
\emph{No evidence} means that no matching read was established, not that a file was unused.
Empty code and missing output are reported separately.
The explicit-unit percentage is the number of required units with explicit read evidence divided by 311.
Every required wildcard remains unresolved for a claim that all its members were used, even when matching member files are read.
Dynamic family references do not count as additional covered inputs.

\section{Cross-Backbone Reuse Experiment (RQ2)}
\label{app:ontology-transfer}

The experiment in Section~\ref{sec:ontology-transfer} evaluates whether an ontology can support task execution with a different generation backbone from the model used to construct it.
We construct the ontology with DeepSeek-V4.1-Flash and reuse it in \ours{} on KramaBench with four generation backbones: Gemma-4-31B-IT, GPT-OSS-120B, DeepSeek-V4.1-Flash, and MiMo-V2.6-Pro.
All generation backbones in this experiment run with thinking disabled, except GPT-OSS-120B, which uses low thinking effort.
We compare \ours{} with smolagents and EvoOntology~\citep{chong2026evoontologyselfevolvingontologylayer} using the KramaBench answer score described in Section~\ref{sec:setup}.
Figure~\ref{fig:answer-scores-backbones} reports the results for each backbone.

\paragraph{EvoOntology configuration.}
We use DeepSeek-V4.1-Flash to construct the EvoOntology baseline. For each KramaBench data lake, DeepSeek-V4.1-Flash synthesizes 60 questions: 40 for initial ontology construction and 20 for ontology evolution. EvoOntology uses agent traces from these synthetic questions to construct and evolve its ontology. 

\section{Component Ablation Details (RQ3)}
\label{app:ablation}

\paragraph{Experimental design.}
The ablations in Section~\ref{sec:ablation} examine the contributions of the ontology-grounded agent, verifier, semantic layer, and experience on KramaBench with DeepSeek-V4.1-Flash. We distinguish changes to ontology construction from changes to the task-solving agent and the ontology information available during execution. The agent and semantic-layer ablations operate on saved ontologies, whereas the verifier and experience ablations modify ontology construction. The verifier ablation accepts every proposed claim, while the experience ablation removes the distiller, so no experience is produced or available in either stage.

\paragraph{Without the ontology-grounded agent.}
We replace the ontology-grounded task-solving agent with smolagents~\citep{smolagents}, while retaining access to the saved ontology. The replacement agent receives an initial ontology retrieval for the task question and can subsequently invoke \texttt{search\_ontology} and \texttt{get\_schema}. These tools provide read-only access to Concepts, Schemas, Fields, Tables, their relationships, and associated experience. Retrieved file references are mapped to the staged benchmark inputs used for execution.

The replacement agent executes Python code over these inputs without the full system's \texttt{ground} interface or its requirement to ground relevant objects before reading lake files. In the full system, grounding adds selected ontology objects to a task graph and determines which lake files the agent can read; the graph also records dependencies between input files, derived artifacts, and the answer. Thus, this variant evaluates the replacement of the ontology-grounded execution procedure while preserving ontology retrieval.

\paragraph{Without the verifier.}
We remove the verifier during ontology construction and automatically accept every claim proposed by the proposer. For each set of proposed claims $H$, the accepted set is $H_{\mathrm{acc}}=H$, and the corresponding changes are applied to the ontology without verifier judgment or verifier-guided revision. The resulting ontology is used by the ontology-grounded task-solving agent. This intervention evaluates the contribution of verification to ontology construction and downstream task performance.

\paragraph{Without experience.}
We remove the distiller during ontology construction, so no experience is produced ($\mathcal{E}=\emptyset$). Neither the construction agents nor the task-solving agent can access experience. Schema induction, semantic enrichment, and verification remain active, and the resulting ontology is used for task execution. The agent can still retrieve Concepts and Links, inspect schemas, ground relevant objects, and compute over the corresponding files using the full system's execution procedure.

This intervention measures the contribution of producing and reusing distilled lessons across both ontology construction and task execution.

\section{Case Study}
\label{app:case-study}

\begin{figure*}[t]
\centering
\footnotesize
\definecolor{casebluedark}{RGB}{38,94,150}
\newcommand{\casestep}[1]{\textbf{\color{casebluedark}#1}}
\begin{tcolorbox}[enhanced, colback=oursblue, colframe=casebluedark, colbacktitle=casebluedark, coltitle=white,
  boxrule=0.6pt, arc=1.5mm, left=2mm, right=2mm, top=1mm, bottom=1mm,
  fonttitle=\bfseries\small, title={(a) Ontology construction on the archeology lake (DeepSeek-V4.1-Flash)}]
\begin{tabularx}{\linewidth}{@{}p{0.15\linewidth}>{\raggedright\arraybackslash}X@{}}
\casestep{Proposer} & Inspects \texttt{climateMeasurements.xlsx} with \texttt{pandas}, which renames repeated headers to \texttt{Age\_ky.1}, \texttt{Age\_ky.2}, \texttt{Age\_ky.3}. Claim \texttt{claim-000001} proposes schema \texttt{climate\_measurements} (Sheet1, header row 5, records 6--8365) with 24 fields: \texttt{Site}, \ldots, \texttt{Age\_ky}, \texttt{Al}, \ldots, \texttt{Ba}, \texttt{PC1}, \texttt{PC2}, \texttt{ODP 967 Dust proxy}, \texttt{ODP 967 wet-dry index}. \\[2pt]
\casestep{Verifier} & Reads the raw header with \texttt{openpyxl}. Columns 19--29: \texttt{Ba}, \textit{blank}, \texttt{Age\_ky}, \texttt{PC1}, \texttt{PC2}, \textit{blank}, \texttt{Age\_ky}, \texttt{ODP 967 Dust proxy}, \textit{blank}, \texttt{Age\_ky}, \texttt{ODP 967 wet-dry index}. \\
 & Verdict: \textsc{refuted}. ``\ldots the inferred Schema is incomplete: the header has 30 columns including three blank/unnamed separator columns (idx 20,24,27) and three additional duplicate `Age\_ky' columns (idx 21,25,28) that the claim does not list.'' No schema is added; the file stays unresolved. \\[2pt]
\casestep{Distiller} & Lesson \texttt{experience-merge-0001-0000} (data quirk), anchored to \texttt{climateMeasurements.xlsx} and \texttt{climate\_measurements.Age\_ky}: ``\ldots three extra columns also carry the header `Age\_ky', so the field Age\_ky corresponds to several physical columns sharing one name. Any analysis keyed on the Age\_ky header may match more than one column.''
\end{tabularx}
\end{tcolorbox}

\vspace{2pt}
\begin{tcolorbox}[enhanced, colback=oursblue, colframe=casebluedark, colbacktitle=casebluedark, coltitle=white,
  boxrule=0.6pt, arc=1.5mm, left=2mm, right=2mm, top=1mm, bottom=1mm,
  fonttitle=\bfseries\small, title={(b) Ontology-grounded execution of task \texttt{archeology-easy-4} (DeepSeek-V4.1-Flash)}]
\textit{Question:} ``What is the amount of calcium in ppm when the dust proxy was the smallest, breaking ties by the minimum wet-dry index? Round your answer to 4 decimal places.''\\[3pt]
\begin{tabularx}{\linewidth}{@{}p{0.15\linewidth}>{\raggedright\arraybackslash}X@{}}
\casestep{Steps 1--2} & \texttt{search\_ontology} for ``dust proxy'', ``calcium ppm'', ``wet-dry index'', ``calcium'', ``dust'', \ldots{} $\rightarrow$ no matching elements. \\
\casestep{Steps 3--4} & Lists the lake root $\rightarrow$ five files, including \texttt{climateMeasurements.xlsx}. \\
\casestep{Step 5} & \texttt{ground("climateMeasurements.xlsx")} $\rightarrow$ file readable (\texttt{in\_ontology: false}); returns the two lessons anchored to the file, including the lesson in (a). \\
\casestep{Step 6} & Reads rows 0--9 of Sheet1 and locates the header at row 5. \\
\casestep{Step 7} & ``There are separator columns and repeated Age\_ky columns.'' Selects columns 12 (\texttt{Ca}), 26 (dust proxy), and 28 (taken as the wet-dry index). The minimum dust proxy, 0.001, occurs in two rows; tie-break values 2688.5 and 2689.0 are ages $\rightarrow$ Ca $=$ 197141.4316 ($\times$). \\
\casestep{Step 8} & ``the wet-dry index column picked was actually `Age\_ky' (header index 28 is Age\_ky, not wet-dry index)''. Prints all 30 header indices. \\
\casestep{Step 9} & Selects columns 12, 26, and 29. Tie-break values 0.3968 and 0.0599 $\rightarrow$ Ca $=$ 207878.8774. \\
\casestep{Steps 10--11} & Confirms a single worksheet and the raw cell values; \texttt{final\_answer(207878.8774)} $\rightarrow$ exact match ($\checkmark$).
\end{tabularx}
\end{tcolorbox}
\caption{\textbf{Case study from saved traces.} (a) The verifier refutes a schema claim that omits repeated \texttt{Age\_ky} columns, and the distiller records the finding as a lesson. (b) Grounding the unresolved file returns that lesson; the agent corrects an initial selection of an \texttt{Age\_ky} column and answers correctly. Steps are condensed; quoted text is verbatim from the traces.}
\label{fig:case-study}
\end{figure*}

Figure~\ref{fig:case-study} traces task \texttt{archeology-easy-4} of KramaBench through ontology construction and task execution, both with DeepSeek-V4.1-Flash. The archeology lake contains five files. The task requires one of them, \texttt{climateMeasurements.xlsx}, a sediment-core record whose single worksheet places several proxy series side by side. Each series block carries its own \texttt{Age\_ky} column, and blank columns separate the blocks. The execution took 11 steps and 59.6K tokens.

\paragraph{Verification keeps a misleading schema out of the ontology.}
The proposer inspected the file with \texttt{pandas}, which silently renames repeated headers, and proposed a schema with a single \texttt{Age\_ky} field. The verifier read the raw header and found the three repeated columns and three separators that the claim omitted. It refuted the claim, the only refutation among the 119 claims for this lake. Had the claim been accepted, the ontology would have described a 24-column layout with one \texttt{Age\_ky} column, the same misreading the agent made in step 7. Construction instead left the file unresolved, so the ontology made no statement about its columns.

\paragraph{Experience retains what a refuted claim revealed.}
A refuted claim adds nothing to the ontology, but the evidence behind the verdict remains useful. The distiller turned the verifier's finding into a lesson anchored to the file and its \texttt{Age\_ky} field. During execution, the lesson reached the agent through grounding (Algorithm~\ref{alg:grounded-action}): it was returned in step 5, when the file entered the task graph. The agent's reasoning in step 7 mentions the repeated \texttt{Age\_ky} columns, and in step 8 it recognized that its chosen column was one of them and re-indexed the header. The trace does not establish that the lesson alone caused this correction, since the agent had also printed the header in step 6. The correction nonetheless follows the layout that the lesson describes, and without it the answer would have been 197141.4316.

\paragraph{Grounding covers files outside the ontology.}
Because no schema described the file, the agent's searches in steps 1--2 matched no ontology element, and the agent listed the lake directly. It then grounded the file by its path. With no schema instantiated by the file, the inspection condition in Eq.~\eqref{eq:schema-before-grounding} holds trivially, and the file became readable. All subsequent reads stayed within the task graph. The case also exposes a cost of conservative construction: an unresolved file is invisible to ontology search, so the agent spent four steps locating it. Retaining such a file as a partially described schema with an attached caveat could shorten this search, which we leave for future work.

\section{Error Analysis}
\label{app:error-analysis}

We analyze the execution trajectories of \ours{} on KramaBench and DiscoveryBench for questions that receive less than full credit. We assign each question one primary error type, summarized in Figure~\ref{fig:error-types}. The seven categories distinguish errors in substance from cases where the answer is right or nearly right in substance but loses credit because of its presentation, scope, or evaluation.

Task misinterpretation leads on KramaBench (32\%), where substantive errors account for 68\% of failures. DiscoveryBench is dominated by evaluation mismatch (40\%) and answer format / scope (29\%), highlighting hypothesis scope and reference matching as its main sources of lost credit.

\noindent\begin{minipage}{\linewidth}
\centering
\includegraphics[width=\linewidth]{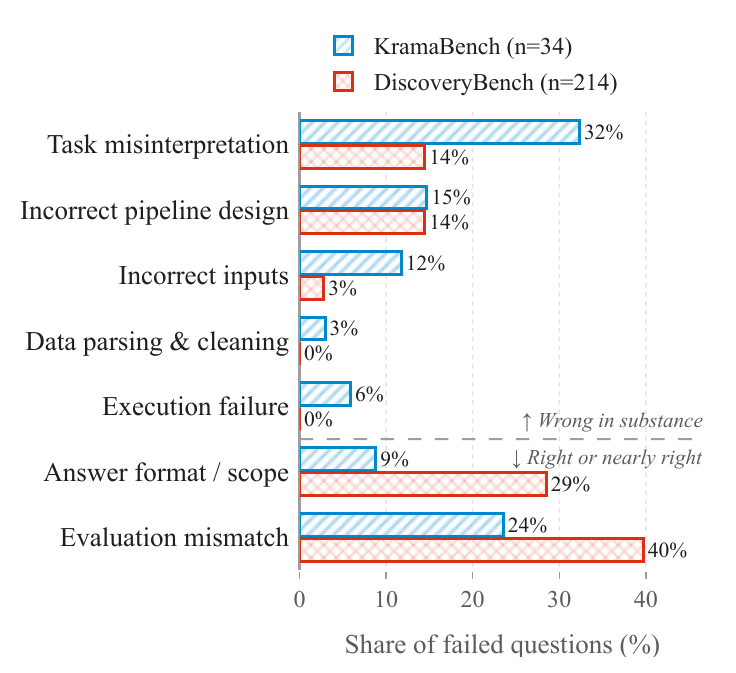}
\captionof{figure}{Primary error types identified from \ours{} execution trajectories on KramaBench and DiscoveryBench. Percentages are computed over questions receiving less than full credit in each benchmark. The dashed line separates errors in substance from cases with right or nearly right substance.}
\label{fig:error-types}
\end{minipage}

\paragraph{Wrong in substance.}
The first five categories identify failures in interpreting or carrying out the analysis:
\begin{enumerate}
    \item \textbf{Task misinterpretation.} The agent answers a different question by using an unintended population, subset, time period, definition, or target quantity. Its answer may be internally consistent but does not address the intended task.
    \item \textbf{Incorrect pipeline design.} The agent understands the question and uses the correct data but constructs the analysis incorrectly. Examples include an incorrect join, filter, grouping key, aggregation, interpolation, time window, order of operations, or statistical test or model.
    \item \textbf{Incorrect inputs.} The agent selects the wrong file, table, sheet, column, or variable, or omits a required source. This category also includes cases where a required file is absent from the data lake and the agent substitutes a proxy.
    \item \textbf{Data parsing \& cleaning.} The source and analysis plan are correct, but loading or cleaning introduces errors, such as mishandled headers, skipped rows, missing values, type or unit conversions, string normalization, duplicates, or date parsing.
    \item \textbf{Execution failure.} The agent produces no answer supported by computation because it crashes, exhausts its step or time budget, repeatedly fails to execute code, or ends with a guess.
\end{enumerate}

\paragraph{Right or nearly right in substance.}
The remaining two categories distinguish output mismatches from evaluation discrepancies:
\begin{enumerate}
    \setcounter{enumi}{5}
    \item \textbf{Answer format / scope.} The answer is right or nearly right in substance but does not match the requested output. On KramaBench, this includes incorrect units, rounding, or representation. On DiscoveryBench, the hypothesis may be too broad or too narrow, introduce unrequested claims or variables, or express its context differently from the reference, preventing the evaluator from matching it.
    \item \textbf{Evaluation mismatch.} The answer is correct or clearly defensible, but credit is lost because of the evaluator or reference. Examples include an incorrect judgment, reference claims beyond the question's scope, a debatable convention in the gold answer, or a gold answer that contradicts the data or its reference pipeline. We assign this label only when supported by concrete evidence; if the agent also makes a substantive error, we assign the corresponding agent-error category instead.
\end{enumerate}

\section{Cost Analysis}
\label{app:cost}

\begin{table*}[ht]
\centering
\small
\setlength{\tabcolsep}{5pt}
\renewcommand{\arraystretch}{1.15}
\begin{tabularx}{\textwidth}{@{}lXrrrr@{}}
\toprule
\textbf{Model} & \textbf{Method} & \shortstack{\textbf{Generation}\\\textbf{/ task}} & \shortstack{\textbf{Construction}\\\textbf{total}} & \shortstack{\textbf{Construction}\\\textbf{/ task}} & \shortstack{\textbf{Total}\\\textbf{/ task}} \\
\midrule
\multirow{6}{*}{DeepSeek-V4.1-Flash} & smolagents & 0.01467 & 0.0000 & 0.00000 & 0.01467 \\
 & Reflexion & 0.03307 & 0.0000 & 0.00000 & 0.03307 \\
 & DS-STAR & 0.02527 & 0.0000 & 0.00000 & 0.02527 \\
 & Data Interpreter & 0.00535 & 0.0000 & 0.00000 & 0.00535 \\
 & \ours{} & 0.01242 & 4.2127 & 0.01228 & 0.02470 \\
 & DatawiseAgent & 0.00936 & 0.0000 & 0.00000 & 0.00936 \\
\midrule
\multirow{6}{*}{GPT-6-Luna} & smolagents & 0.01300 & 0.0000 & 0.00000 & 0.01300 \\
 & Reflexion & 0.02807 & 0.0000 & 0.00000 & 0.02807 \\
 & DS-STAR & 0.03219 & 0.0000 & 0.00000 & 0.03219 \\
 & Data Interpreter & 0.00245 & 0.0000 & 0.00000 & 0.00245 \\
 & \ours{} & 0.00415 & 0.5053 & 0.00147 & 0.00562 \\
 & DatawiseAgent & 0.00581 & 0.0000 & 0.00000 & 0.00581 \\
\midrule
\multirow{6}{*}{MiMo-V2.6-Pro} & smolagents & 0.02394 & 0.0000 & 0.00000 & 0.02394 \\
 & Reflexion &  0.04867 & 0.0000 & 0.00000 &  0.04867 \\
 & DS-STAR & 0.06265 & 0.0000 & 0.00000 & 0.06265 \\
 & Data Interpreter & 0.00813 & 0.0000 & 0.00000 & 0.00813 \\
 & \ours{} & 0.03024 & 9.7268 & 0.02836 & 0.05860 \\
 & DatawiseAgent & 0.02859 & 0.0000 & 0.00000 & 0.02859 \\
\bottomrule
\end{tabularx}
\caption{Generation and ontology construction costs in USD. Per-task costs use 343 runs per row; construction costs are amortized over those runs. Per-task values are rounded to five decimal places, with lower bounds rounded down.}
\label{tab:cost-per-task}
\end{table*}

Table~\ref{tab:cost-per-task} reports costs in USD over 343 selected task runs for each model--method pair. Generation cost per task is the reported generation total divided by 343. We report ontology construction both as a total and as an amortized cost per task, also divided by 343; total cost per task includes both components.
\raggedbottom
\section{Agent Prompts}
\label{app:prompts}

The following boxes present condensed versions of the verifier, proposer, and task-execution prompts. We use File uniformly for a source file, worksheet, or record region, identified by its source path and any applicable sheet or row range.

\begin{agentprompt}{Verifier Prompt}
Independently verify outstanding Claims about one Lake's Ontology against relevant Lake Files and accepted context in the JSON task. Claims are proposals, not evidence; they may reference objects added by other Claims, and context may reflect completed verdicts. Do not rewrite Claims or use Proposer reasoning.

Check inferred Schemas and unsupported File-to-Schema assignments. Do not reinterpret declared Schema meanings supplied by Documentation.
\end{agentprompt}

\begin{agentprompt}{Proposer Prompt}
Build one Lake's Ontology from JSON stage, accepted context, and remaining_work. Read context once into Python. Repropose revisions with the same id. Leave Experience to its writer.

FILE_MAPPING
- Inspect Documentation first: files gives absolute paths; documentation_candidates are unverified filename hints. Reuse accepted definitions; otherwise consult known Documentation as support, not assigned data.
- Inspect every assigned File's record regions and nonempty worksheets once; reuse observations; Propose only Schema and File-to-Schema assignment Claims; no removals.
- Declared Schemas cite existing Documentation and copy its Field meanings, never inferred meanings. Documentation is support, not data. Mapped Files require structural checks and accepted Schemas.

SCHEMA_ENRICHMENT
- Use supplied schema_ids, accepted summaries, Concepts, Links, and get_schema definitions to propose Concepts/Links, including shared meanings across Schemas. Revise/remove only Concepts/Links; preserve Schemas, Fields, and Files.
- Never reread raw Files, rediscover structure, or revalidate documented meanings. Record specific evidence gaps in unresolved; revisit Schemas/remaining_work, not completed Files.
- Submit small useful batches; skip unchanged objects and exact refuted/undecidable entries in decided_proposals. Empty claims are valid when no supported changes remain.

WORKFLOW
- get_schema(schema_id) returns Fields/Files; search_ontology(query, hops=2) returns matches/neighbours. Never search empty context. Only submitted, accepted Claims update the Ontology; rereading cannot expose pending Claims.
- Batch related Python steps/tool calls with known formats; reuse metadata, cache intermediates, and print only counts, relevant samples, or problems.
\end{agentprompt}

\begin{agentprompt}{Task-Execution Prompt}
Answer a Lake task by finding data in the Harness, grounding it in a task graph, and computing with Python. The JSON task supplies task (question and optional answer_format), max_steps (step budget), and workspace.

Use File for a source addressed by its path and, when applicable, worksheet or record region. Apply the same search, grounding, and access rules to these sources.

TOOLS
- search_ontology(query, hops=2) returns matched objects and neighbours within hops edges: Concepts, Fields with schema_id, Schema summaries, Files, traversed edges (source, kind, target), paths from matches to neighbours, and Experience. Files omitted from a Schema's listing are partitions of the same dataset; search path terms to find them.
- get_schema(schema_id) returns a Schema, all its Fields, and its first associated Files.
- ground(ids) accepts one id or a list: Schema (with all associated Files), File, Field, Concept (with Links and connected Fields, Files, and Concepts), or Lake path. A path resolves to its mapped sources. It adds them to the task graph and returns additions, Schema Fields, edges, newly readable Files, and relevant Experience.

FILES
- LAKE_ROOT is read-only. Read only grounded Files or Documentation of grounded Schemas. Otherwise, PermissionError requires grounding the File before retrying.
- Write created, converted, cleaned, joined, or split files in WORKSPACE; each becomes a task-graph node.

WORKFLOW
1. Preserve the question's population, filters, measure, period, and units exactly.
2. Search, ground the next computation's data, and compute. These may share one step; ground more as needed. Inspect associated Schema descriptions before grounding mapped Files.
3. Call final_answer as soon as answered. Print only counts, small samples, and computed values.
\end{agentprompt} 

\end{document}